\documentclass{article}

\usepackage{arxiv}

\usepackage[utf8]{inputenc} 
\usepackage[T1]{fontenc}    
\usepackage{hyperref}       
\usepackage{url}            
\usepackage{booktabs}       
\usepackage{amsfonts}       
\usepackage{nicefrac}       
\usepackage{microtype}      
\usepackage{lipsum}		
\usepackage{graphicx}
\usepackage{natbib}
\usepackage{doi}
\usepackage{amsmath} 
\title{Grammatical Structure and Grammatical Variations in Non-Metric Iranian Classical Music}

\author{\small
  \begin{tabular}{@{}c@{}c@{}c@{}}
    \begin{minipage}[t]{0.32\textwidth}
      \centering
      Maziar Kanani \\
      School of Computer Science \\
      University of Galway \\
      Galway, Ireland \\
      \texttt{m.kanani1@universityofgalway.ie}
    \end{minipage} &
    \begin{minipage}[t]{0.32\textwidth}
      \centering
      Seán O'Leary \\
      School of Computer Science \\
      TU Dublin \\
      Dublin, Ireland \\
      \texttt{sean.oleary@tudublin.ie}
    \end{minipage} &
    \begin{minipage}[t]{0.32\textwidth}
      \centering
      James McDermott \\
      School of Computer Science \\
      University of Galway \\
      Galway, Ireland \\
      \texttt{james.mcdermott@universityofgalway.ie}
    \end{minipage}
  \end{tabular}
}

\date{}
\begin{document}

\maketitle

\begin{abstract}

In this study we introduce a symbolic dataset composed of non-metric Iranian classical music, and algorithms for structural parsing of this music, and generation of variations.

The corpus comprises MIDI files and data sheets of Dastgah Shour from Radif Mirza Abdollah, the foundational repertoire of Iranian classical music.

Furthermore, we apply our previously-introduced algorithm for parsing melodic structure ~\citep{kanani2023parsing}to the dataset. Unlike much Western music, this type of non-metric music does not follow bar-centric organisation. The non-metric organisation can be captured well by our parsing algorithm. We parse each tune (Gusheh) into a grammar to identify motifs and phrases. These grammar representations can be useful for educational and ethnomusicological purposes.

We also further develop a previously-introduced method of creating melodic variations~\citep{kanani2023parsing}. After parsing an existing tune to produce a grammar, by applying mutations to this grammar, we generate a new grammar. Expanding this new version yields a variation of the original tune. Variations are assessed by a domain-expert listener. Additionally, we conduct a statistical analysis of mutation with different representation setups for our parsing and generation algorithms. The overarching conclusion is that the system successfully produces acceptable variations post-mutation. While our case study focuses on Iranian classical music, the methodology can be adapted for Arabic or Turkish classical music.

\end{abstract}

\section{Introduction}

A substantial portion of the classical music culture in Iran, Turkey, Arabic regions, and parts of India is characterized by non-metric or free rhythm music. Unlike \textit{Alap} in Indian music or \textit{Taqsim} in Arab and \textit{Makam} in Turkish music, the main body of Iranian classical music, \textit{Avaz}, is non-metric~\citep{tsuge1970rhythmic,kanani2019role}. This distinctive feature prompted us to focus on Iranian classical music as a case study for our exploration of musical structure and structural mutation in non-metric musical cultures. There exists a structured repertoire within this musical tradition called Radif, which serves as a pedagogical framework for learning improvisation through various Dastgahs (musical systems) and Gushehs (melodies within these systems). The pieces in Radif are divided to 12 subsets: 7 Dastgahs and 5 Avazes, each comprising numerous smaller music pieces referred to as Gushehs. The term Avaz has three connotations: broadly, it signifies singing; it denotes the segments of Radif smaller than a Dastgah; and specifically, it refers to the non-metric Iranian music style. In this study, our focus is on the latter meaning of Avaz, excluding its associations with singing or smaller Dastgahs.

In this study we introduce the first symbolic dataset of such music.

This type of music necessitates a flexible micro-level analysis rather than a bar-centred analysis which can be suitable for Western compositions~\citep{doherty2022melodic}. Avaz generally avoids the formal structures (such as Sonata, Theme and Variation, or Rondo Forms) prevalent in Western music~\citep{farhat1978form}, often initiating with a musical idea that is developed sequentially without adhering to predefined forms.

Gushehs in Iranian classical music are divided into three types: modal, melodic, and rhythmic. Melodic Gushehs, like \textit{Hazin}, and rhythmic ones, such as \textit{Kereshmeh}, have distinct melodies or rhythms that are fixed and cannot be altered. This means variations on these Gushehs may not be acceptable as examples of the same Gusheh.

Radif has been taught by various artists over the last century. The most renowned version is Radif Mirza Abdollah. We've chosen to use a recent edition of this Radif by Dariush Talai~\citep{talai2015} for our study because it includes both a transcription and an explanation of the music's hierarchical structure.

Representing this music accurately involves tackling two main challenges: non-metric rhythm and micro-tonal pitch. These challenges are addressed in Sec.~\ref{sec:corp}.

The sequential melodic data for each piece in the collection is parsed by a previously-introduced algorithm~\ref{sec:corp}, extended to be suitable for Iranian classical music. The result is a grammar which represents the hierarchical repetition structure of the piece, which is potentially valuable for educational and ethnomusicological research. 

We then also applied mutations to these grammars, in order to generate variations of the piece which respect overall structural properties, such as complexity and reuse. It preserves the hierarchical structure while introducing significant variations compared to the original tune.

We are not looking to generate new music but rather variations. This approach to generating variations is useful and exciting for musicians, researchers, and listeners. For instance, a musician with several variations of a Gusheh can improve their improvisation skills in this style of music. Researchers can benefit when they need several versions of a Gusheh but lack enough real recorded samples. Finally, listeners can use the algorithm to obtain variations of the tunes they like, providing them with new versions of their favorite music. Additionally, this avenue of study provides some evidence that our approach to representing hierarchical structure is valid.

 Above, we created variations at the Gusheh (piece) level. Playing all of the Gushehs in order simulates a Dastgah performance, which is traditionally taught to musicians. The next step for a music student is to create variations of a Dastgah for their own performances. This gives us the motivation to also try our mutations on the entire Dastgah to observe the variation results.

\section{Related Works}

 In~\citep{farhat1978form} Farhat has a wide research on form and style in Iranian classical music. He traces the classical form back to ancient times, noting its separation from the broader Islamic musical tradition in the 16th century. This shift gave Persian music its unique character, despite retaining basic similarities with Turkish and Arabic music. He outlines the Dastgah system and modal structures~\citep{farhat1978form, farhat2004dastgah} and compares form in western music and Iranian music. He believes Persian music's expressiveness and flexibility allow it to be shaped by the individual tastes and temperaments of its performers.

Several authors have discussed Iranian non-metric music. Tsuge explains the concept of non-metric and how it is more important in Iranian music than others~\citep{tsuge1970rhythmic}. He notes that the rhythmic organization of Avaz primarily relies on the poetic meter system, with a recurring cycle of short and long syllables forming the basis of its rhythmic structure. This structure is deeply bound up with the nature of the Persian (Farsi) language and its classical versification system, which plays a significant role in how the melody is formed and perceived. In~\citep{kanani2020study} the authors introduce the main melodic figures that are usually used in non-metric Avaz music. 

Laudan Nooshin broadly discussed improvisation in Iranian and Indian music~\citep{nooshin2006improvisation}. Her work on improvisation is mostly constructed on how improvisation is shaped in non-metric Iranian music based on Radif by different prominent performers. In another place she explores what creativity means in this music~\citep{nooshin2017iranian}. Jean During has done substantial work in Middle Eastern music~\citep{during1997systeme, during2006intervals, during1991art} and his role is very important as the first published transcription of Radif Mirza Abdollah was by him. However we used~\citep{talai2015} by Dariush Talai for our work because he provided a more understandable explanation of Radif for readers.

At the start of the study we tried to find a symbolic corpus that covers middle eastern classical music but we didn't find any. Turkish Makams have a symbolic corpus called SymbTr~\citep{karaosmanouglu2012turkish}. According to our knowledge there is no symbolic corpus for Iranian and Arabic music. However there were some audio datasets for these types of music like {\em Dunya} corpus which includes Turkish Makam, Carnatic, Hindustani, Beijing Opera, Arab-Andalusian. Its Turkish part is connected to our research area~\citep{uyar2014corpus}. KUG Dastgāhi~\citep{nikzat2022kdc} and~\citep{heydarian2005database} are two audio corpora for Iranian music. Nava~\citep{baba2019nava} is an audio dataset for Iranian instrument recognition task and Ar-MGC: Arabic Music Genre Classification Dataset~\citep{almazaydeh2022arabic}. Therefore, we have decided to create our own dataset representing a subset (Dastgah Shour) of the repertoire (Radif), for use in our algorithms and for use by others. In the future, we plan to expand the dataset.

Hakkı Parlak and Kösemen discuss~\citep{parlak2018automatic} an approach to automatic music generation for Turkish Makams using random numbers. The system utilizes these random numbers to determine the notes, note lengths, tempo, instruments, and percussion for compositions within chosen Turkish Makams. While the system aims to generate music that can be completely random or conform to musical rules based on user preferences, it does not seek to replicate the works of past composers but rather to introduce new musical ideas and provide entertainment and inspiration for enthusiasts of Turkish Makam music.

Şentürk and Chordia~\citep{sertan2011modeling} explore the application of Variable-Length Markov Models (VLMM) to predict melodies in the Uzun Hava form of Turkish folk music. This study introduces the first symbolic machine-readable database of Uzun Havas and represents the first attempt at predictive modeling in Turkish folk music. The research highlights the predictive power of VLMMs in modeling non-Western and non-metric musical styles, suggesting their broader applicability beyond Western music traditions. 

\section{The corpus\texorpdfstring{\protect\footnote{\url{https://github.com/maziarkanani/ShourCorpus}}}{ (see footnote)}}
\label{sec:corp}

\begin{figure}[h]
  \centering

  \makebox[\textwidth][c]{\includegraphics[width=1.3\textwidth, trim={0 1cm 0 1cm}, clip]{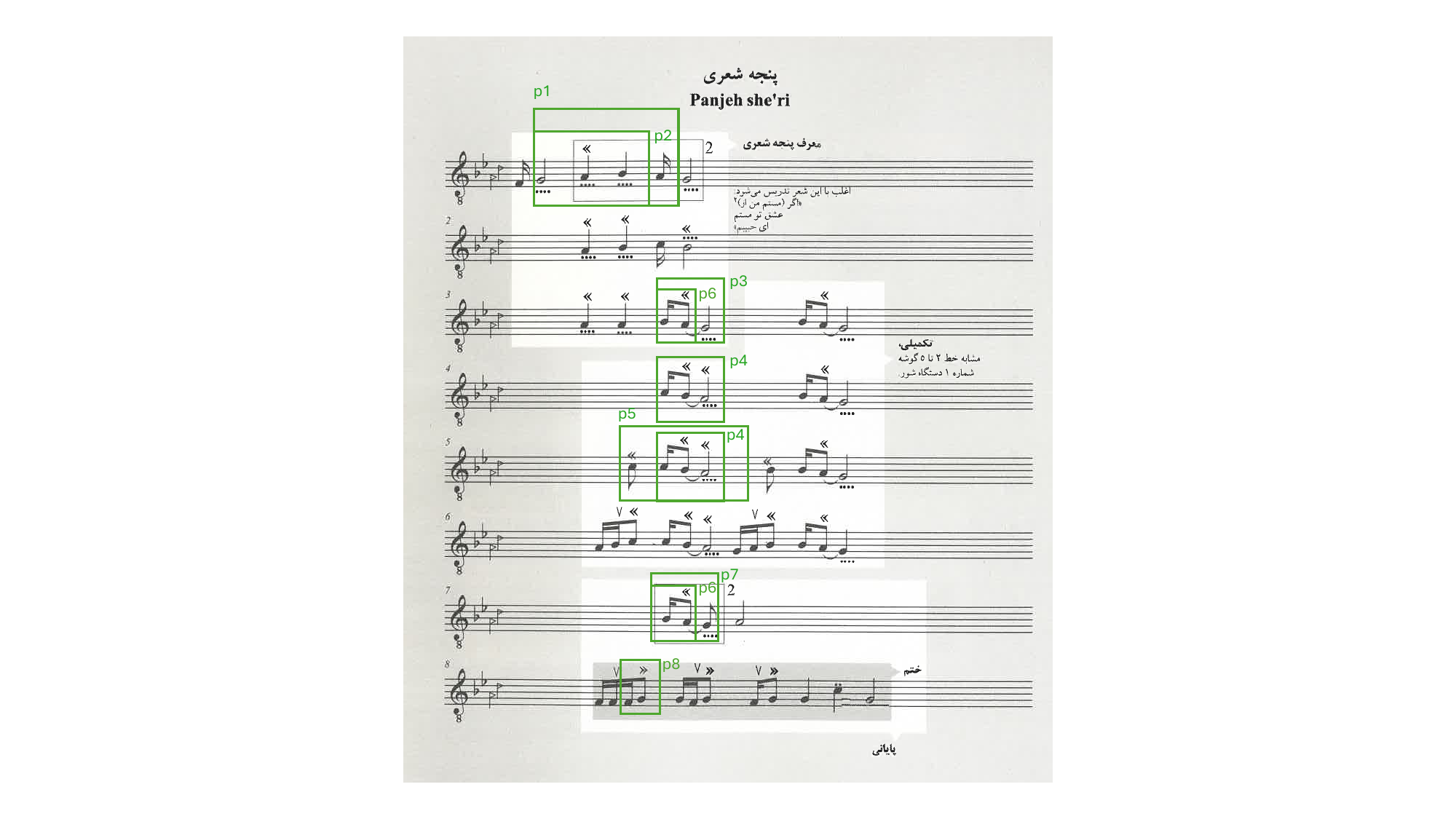}}
  \caption{Transcription sample from Dariush Talai's``Radif Analysis," illustrating the second Gusheh structure in the Shour Dastgah. The three white boxes indicate the main structural divisions, while the last white box contains a further sub-division marked in grey. The green boxes show the first time a rule appears, detected by the algorithm and added by the authors. \label{fig:transcript}}
\end{figure}

In the realm of non-metric music, Radif stands as the foundational repertoire of Iranian classical music. With numerous eminent performers having crafted their interpretations, various versions of Radif exist. However, the version by Mirza Abdollah has emerged as the predominant choice in music education throughout the current century. Among the various transcriptions of Mirza Abdollah's Radif, we have chosen to work with ``Radif Analysis - based on the notation of Mirza Abdollah's Radif with annotated visual description'' by Dariush Talai~\citep{talai2015}. This particular edition offers a detailed hierarchical structure for each Gusheh, accompanied by a recorded performance that aligns with the musical scores and our corpus. Radif is traditionally divided into 12 subsets: Shour, Bayat-e-tork, Dashti, Abu-ata, Afshari, Segah, Nava, Homayoun, Bayat-e-Esfahan, Chahargah, Mahur, and Rast-Panjgah. Our corpus focuses on Shour, one of the primary and most extensive subsets, comprising 29 non-metric tunes (Gushehs). The total number of notes in the MIDI files is 7001, and it takes 1965.81 seconds to be played. Figure~\ref{fig:transcript} presents a transcription example from the book.

\begin{figure}
  \centering
  \includegraphics[width=0.6\textwidth]{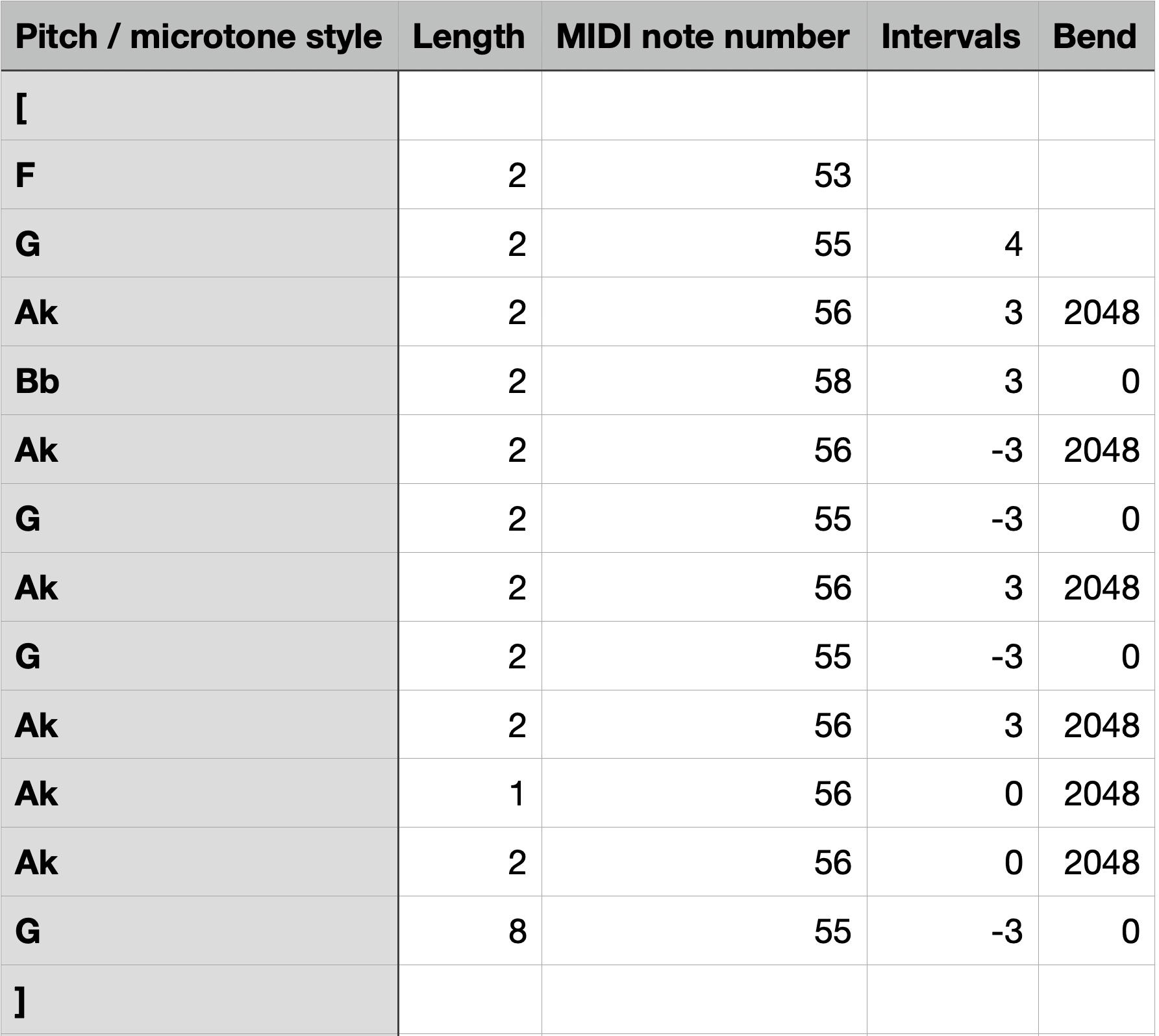}
  \caption{An example of data sheets \label{fig:daramad}}

\end{figure}

In our dataset, we represented each musical piece as a sequence of notes, where for each note we store note name, note duration, MIDI note number, interval, and pitch bending, as illustrated in Figure~\ref{fig:daramad}. 

\subsection{Tonality}
Notes are denoted by the symbols ${C, D, E, F, G, A, B}$, with accidental signs including flat ($b$), koron ($k$), sori ($s$), and sharp ($\text{\#}$). Notably, ``koron'' and ``sori'' indicate micro-tonal adjustments specific to Iranian music - quarter tones lower and higher, respectively.

\subsection{Chromatic Scale}
While these intervals suggest a chromatic scale with 24 quarter notes in one octave, all of which may be used in contemporary compositions, Iranian classical instruments typically play only 18 of these notes: {C, Db, Dk, D, Eb, Ek, E, F, Fs, F$\text{\#}$, Gk, G, Ab, Ak, A, Bb, Bk, B} with quarter-note intervals {2, 1, 1, 2, 1, 1, 2, 1, 1, 1, 2, 1, 1, 2, 1, 1}.

\subsection{MIDI}
We represent quarter-tones using regular MIDI note numbers by performing an adjustment on them. Koron and Sori are two types of quarter-tones, respectively, a quarter lower and a quarter higher than the natural note. To represent Koron, we use a semitone (flat) lower than the natural note, and for Sori, we use the same number as the natural note. Then, we employ pitch-bending to make Koron and Sori distinguishable from flat and natural notes. The bend column is used to represent micro-tonal sounds in MIDI outputs by setting the bend for micro-tones at 2048 units: a flat note is transformed into the note koron variant, and a natural note is altered to its sori. This value is 0 for standard MIDI tones. In our algorithms, we represent pitches using just MIDI pitch numbers. This is sufficient because, similar to Western diatonic music, no Gusheh will include more than one altered note of the same scale degree, e.g.~if D is present then Ds will not be present.


\subsection{Octaves}
The lowest note in the first Gusheh is the start note for the main octave. For example in the first tune of our corpus the lowest note is F3 so the main octave is from F3 to F4. The music in the main octave is represented only by its symbols and accidental signs (if any are present). For notes in octaves other than the main one, we use ‘+’ or ‘-’ followed by a number to indicate the number of octaves above or below the main octave. For example, an Ak from one octave higher than the main octave would be written as Ak+1.

\subsection{Intervals}
The Intervals column quantifies the pitch difference between consecutive notes, where ``1" signifies a quarter-tone. 

\subsection{Durations}
Note durations are inexact in this form of music but can be categorized into four main types: very short, short, long, and very long, corresponding to sixteenth, eighth, quarter, and half notes~\citep{talai2015}, and are represented numerically as 1, 2, 4, and 8.

\subsection{Greater Hierarchical Structures}
We also notated the hierarchical structure of the piece, which is given in the original printed source (see Figure~\ref{fig:transcript}). In our notation, brackets signify hierarchical relationships, giving a tree structure. An open bracket ``['' marks the beginning of a node (a section or subsection), and each close bracket ``]'' signifies the end of that node. In each tune, the entire tune is enclosed between brackets, giving the root node. Every other pair of brackets represents a child node, and a child node can itself have other child nodes (brackets) that represent sections and subsections. Taking Fig.~\ref{fig:transcript} as an example, the abstract structure is \verb+[ [ ] [ ] [ [ ] ] ]+. The outer brackets frame the tune. The second open and close brackets contain the first section that includes the first three lines in Fig.~\ref{fig:transcript}. The third open/close brackets surround the section from line three to 6. The next open bracket is accompanied by another open bracket which means it has a subsection. That is for line 8 and nine and we can see the subsection at the last line highlighted. Due to the fact that these subsections are not repeated in the Gusheh, our algorithm (described below) does not detect them. However, this is not what we are seeking in this paper: our algorithm is focused on detecting repetition structure. Since some of these subsections are repeated in some other Gushehs, running the algorithm over the entire concatenated Radif will detect many of them. The hierarchical structure notation will be used in future research.

\section{Methodology}

\subsection{Representation}

Based on MDL (Minimum Description Length) ideas, the shorter representation of an object captures its internal structure better~\citep{li2008introduction, grunwald2007minimum, schmidhuber1997low}. In our previous work we operationalised this by representing sequences as grammars, as described in Section~\ref{PAI}~\citep{kanani2023parsing}. Thus, we propose multiple representations for musical data motivated by finding a representation which gives short encoding. In this subsection we will see the integer/tuple, chromatic/diatonic and interval/pitch possibilities for the representations.

\subsubsection{Integer/Tuple}
We opted to represent each note as a tuple, including the note's pitch and its duration. In our previous work, we represented notes as integers (MIDI note numbers), because in Irish folk music most notes have the same duration. 

\subsubsection{Chromatic/Diatonic and Pitch/Interval}

Another critical decision in our approach is choosing between pitches or intervals for the representation. The initial preference for a pitch-based representation is that it tends to have a shorter encoding than those based on intervals for the Iranian music corpus. However, given the hypothesis that Iranian music frequently employs repetition with transposition, exploring interval-based approaches seemed promising.

The example shown in Figure~\ref{fig:daramad} depicts a sequence of chromatic MIDI note numbers and their durations: 

\noindent[(53,2), (55,2), (56,2), (58,2), (56,2), (55,2), (56,2), (55,2), (56,1), (56,2), (55,8)]

\noindent The unique set of pitches for the entire tune is \{53, 55, 56, 58, 60\}. When we consider these pitches as degrees within a diatonic modal framework, they can be mapped as \{53:1, 55:2, 56:3, 58:4, 60:5\}. Based on this mapping, the diatonic interval list for the sequence would be represented as:

\noindent[(0,2), (1,2), (1,2), (1,2), (-1,2), (-1,2), (1,2), (-1,2), (1,1), (0,2), (-1,8)] 

\noindent where the first elements reflect the relative pitch change between notes according to their position within the modal framework. As there is no previous note there is no interval there for the first note so we dedicate 0 to the first tuple to make it possible to represent the duration (that is 2 here).

\subsection{Parsing} \label{PAI}
To capture the musical tune structures, we employ the concept of Pathway Assembly (PA)~\citep{marshall2019quantifying}. PA is a method that detects hierarchical repetition structure, in our case in sequences.

Given a sequence such as \verb+abracadabra+, the elementary units are identified as \verb+{a, b, c, d, r}+, and the PA unfolds by iterative concatenation as follows:

\verb|{a, b, c, d, r} -> ca -> ab -> ra -> cad |

\verb|-> abra -> cadabra -> abracadabra|~\citep{marshall2019quantifying}

The PA Index (PAI) quantifies the minimum length of the PA required for any given object, with the PAI for 'abracadabra' determined to be 7, as shown by the 7 concatenation steps, above.

To understand the structure of a piece of music and determine its PAI, we employ the Sequitur algorithm~\citep{nevill1997identifying}. The algorithm identifies hierarchical patterns within data by iteratively substituting consecutive symbol pairs with a new non-terminal symbol. This method of redefining sequences enables a clear representation of the tune's structure, illustrated through the example of \verb|abracadabra| below:

\begin{verbatim}
    p0 -> p1 c a d p1
    p1 -> p3 p2
    p2 -> r a
    p3 -> a b
\end{verbatim}

Although Sequitur rules may combine several items and PA only permits binary concatenation, the two are equivalent as the rules' right-hand sides can be segmented into multiple binary joins. This allows for the PAI calculation, where, for 'abracadabra', the sum of joins within the rules equals 7. The algorithm translates a sequence into tune structures, enabling understanding musical patterns. We can also reverse this process later, after mutation of the grammar, to expand it and produce a sequence that represents our generated tune.

\begin{figure}
    \centering
    \includegraphics[trim=0 100 0 100, clip, width=0.8\textwidth]{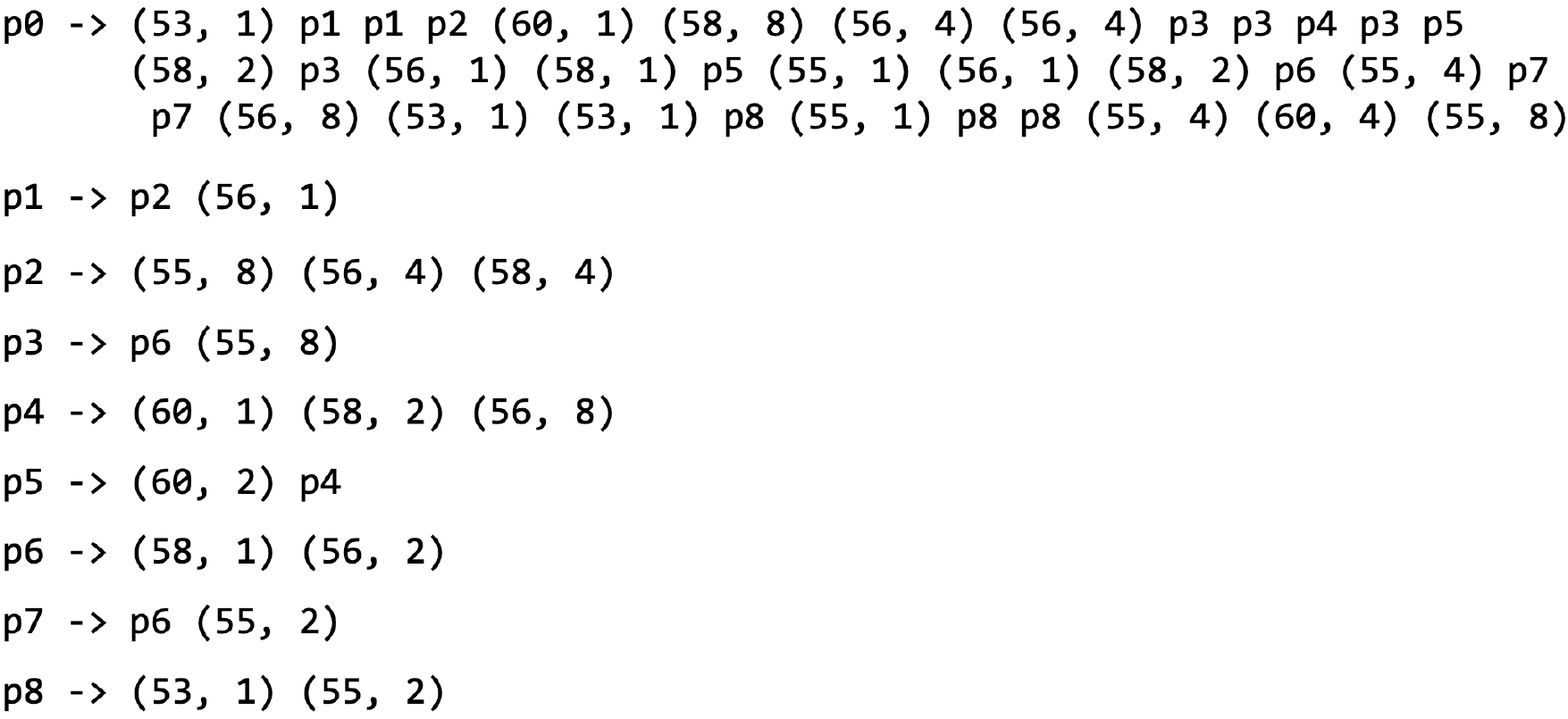}
    \caption{Grammar for the Gusheh shown in Fig.~\ref{fig:daramad}}
    \label{fig:enter-label}
\end{figure}

Fig.~\ref{fig:enter-label} is the grammar based on Fig.~\ref{fig:transcript}. Each note is represented as a tuple: pitch and duration. 

\subsection{Mutations}
\begin{figure}
    \centering
    \includegraphics[trim=0 250 100 0, clip, width=1\textwidth]{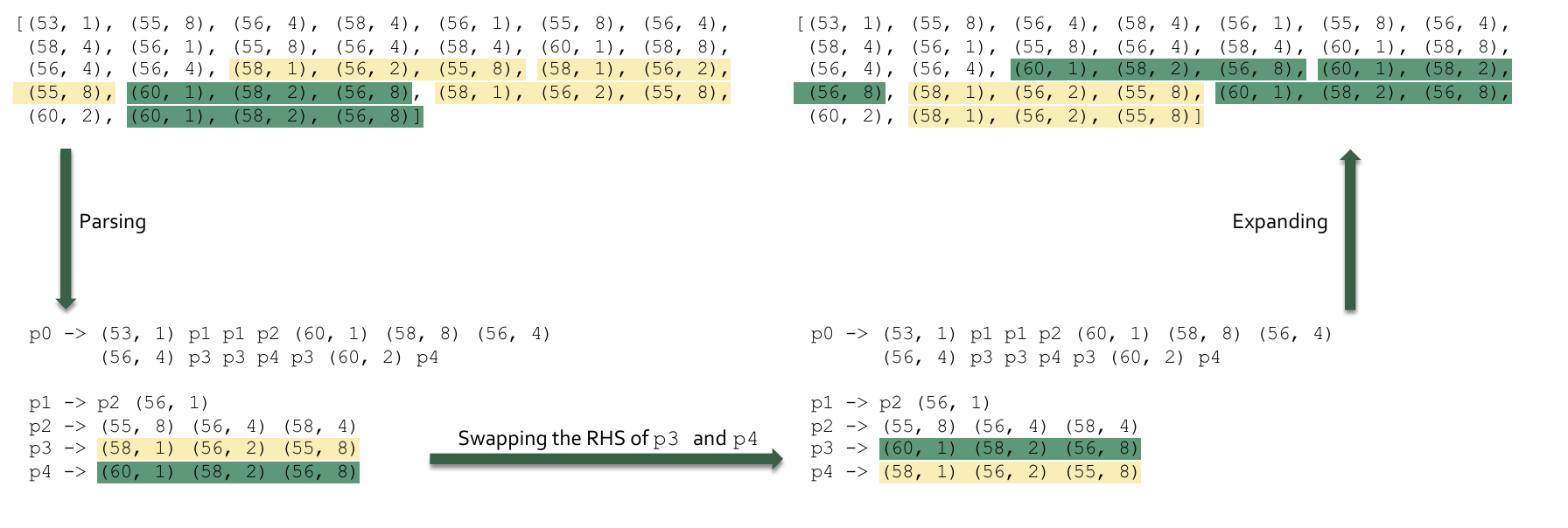}
    \caption{The phrase (58, 1), (56, 2), (55, 8) occurs three times, highlighted yellow in the original tune (top-left). This becomes the RHS of rule p3 in the grammar (bottom-left). The random mutation swaps the RHS of rules p3 and p4, giving a new grammar (bottom-right). Deterministic expansion leads to the variant tune (top-right).}
    \label{fig:mutations}
\end{figure}

In our previous work~\citep{kanani2023parsing,kanani2023graph} we introduce 19 distinct types of mutations (Table\ref{tab:my_label}) that can be applied to the grammatical representation of a tune, encompassing operations such as addition, deletion, swapping, and inversion of grammar parts. These mutations are designed to ensure that the resultant musical variations retain the connections to the original piece, while introducing novel elements that enrich the musical experience. Fig.~\ref{fig:mutations} is an example of how a single mutation can affect a tune.

\begin{table}
    \centering
    \begin{tabular}{|l|l|}
    \hline
    \textbf{Definition} & \textbf{Definition} \\
    \hline
    Insert a primitive in an RHS      &      Delete a primitive from an RHS  \\
    \hline
    Move a primitive – in an RHS      &     Move a primitive – to another RHS  \\
    \hline
    Swap two primitives - in an RHS   &   Swap two primitives – between two RHSs \\
    \hline
    Change a primitive in an RHS      &  \\
    \hline
    \hline
    Insert an existing rule to an RHS &      Delete a rule from an RHS   \\
    \hline
    Move a rule - in an RHS           &     Move a rule – between two RHSs \\
    \hline
    Swap two rules – in a single RHS  &     Swap two rules – between two RHSs \\
    \hline
    Swap a rule and a primitive - in an RHS &   Swap a rule and a primitive - in two RHSs \\
    \hline
    Reverse an RHS                    &     Reverse a sub-sequence of an RHS\\
    \hline
    Swap two RHSs & \\
    \hline
    Delete a rule from the grammar & \\
    \hline
    \end{tabular}
    \caption{Mutation types - the abbreviation ``RHS" stands for ``right-hand side". \label{tab:my_label}}
\end{table}

\subsubsection{Forward/Backward}

If we use intervals for the representation, in the expanding step, a starting point is needed to convert these intervals back into pitches, thus creating the variant tune. This approach draws from the understanding that mutations in intervals might drastically alter the tonality unless carefully managed. One approach could be to retain the first note of the original tune as the starting point and sequentially apply the mutated intervals to derive the subsequent pitches. However, this method results in variations that always begin with the same note, which is not typically the case in natural variations. In practice, variations crafted by musicians often share the same concluding note rather than the starting one. Therefore, There is another possibility to preserve the last note of the tune and run the Sequitur algorithm on the reversed sequence. To this end, we experimented with a interval backward generation setup, ensuring to preserve the tonality by fixing the last note of each sequence. 

\subsubsection{Repair}
To reduce the potential for tonal deviation post-mutation, we adhered to the practice of maintaining the final note of each tune—a common anchor across different versions of a Gusheh—and then reconstructing the melody backwards from this note. This method, however, introduced a new challenge: ensuring the melody remains within the original modal framework. The term ``modal framework" refers to the set of notes that are utilized within a particular section or Gusheh, defining its unique scale or mode~\citep{chalesh2017seyir, Kanani2018}. Repair is applied when a note's pitch falls outside the predefined modal framework's range, defined by a unique list of pitches. We addressed the challenge by adjusting the generated intervals to match an ordered set of pitches present in the original tune, always selecting the closest pitch within the modal framework boundaries. This type of repair is called \textit{clamp}.

There is another type of repair called \textit{mirror}. If a generated note is beyond the highest or lowest pitch in this list, the mirroring method is employed to adjust the note back within the acceptable range. This adjustment is made by calculating the difference between the out-of-range note and the closest boundary (either the highest or lowest pitch). This difference is then applied in the opposite direction from the opposing boundary, effectively ``mirroring'' the note back into the modal framework. This is done at the diatonic level, so that the mirrored note is within the framework.


\subsection{Setup Options}

\begin{figure}
  \centering
  \includegraphics[trim = 0 200 120 180, width=1\textwidth, clip]{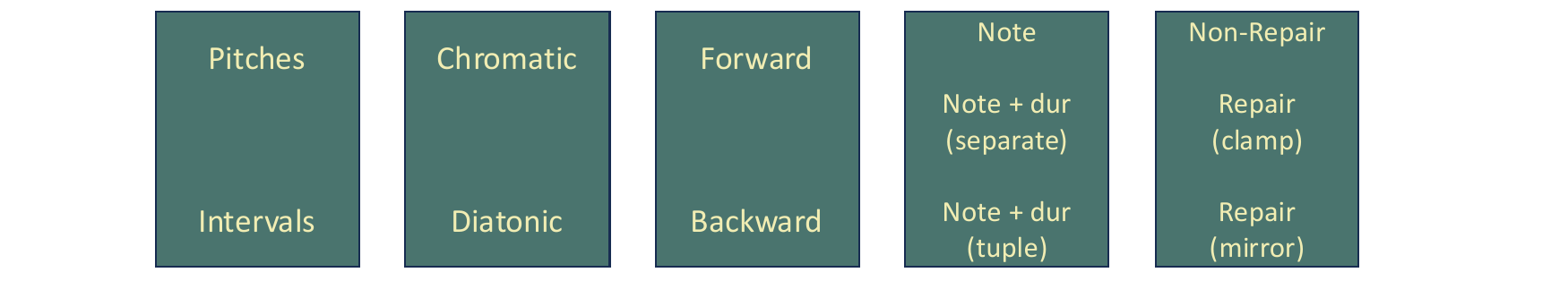}
  \caption{The setup options for the algorithm \label{fig:mind_map}}
\end{figure}

Figure~\ref{fig:mind_map} presents the potential setups that can be achieved by selecting one option from each category of the 5 discussed above. We selected 5 specific setups out of 72 potential setups. Differing from our prior work focused on Irish music, which employed a setup characterized as ``pitches, chromatic, forward, note, non-repair'', our exploration into Iranian music evaluates five distinct setups: one pitch-based and four interval-based.

The pitch-based setup is categorized as ``pitches, chromatic, forward, note + duration (tuple), non-repair''. The first of the interval-based setups is described as ``intervals, chromatic, forward, note + duration (tuple) and clamp''. Second one, ``intervals, chromatic, backward, tuple and clamp''. The third interval-based setup is characterized as ``intervals, diatonic, backward, tuple and clamp''. Finally, the last interval-based setup repeats the configuration of the third but is distinguished by mirror repair instead of clamp.






\section{Pathway Assembly Algorithm Evaluation on Iranian Classical Music\texorpdfstring{\protect\footnote{\url{https://github.com/maziarkanani/Grammatical-Structure-and-Grammatical-Variations-in-Non-Metric-Iranian-Classical-Music/tree/main}}}{ (see footnote)}}
\label{sec:Seq}

\begin{figure}[ht]
    \centering
    \begin{minipage}{0.46\textwidth}
        \centering
        \includegraphics[width=1\linewidth]{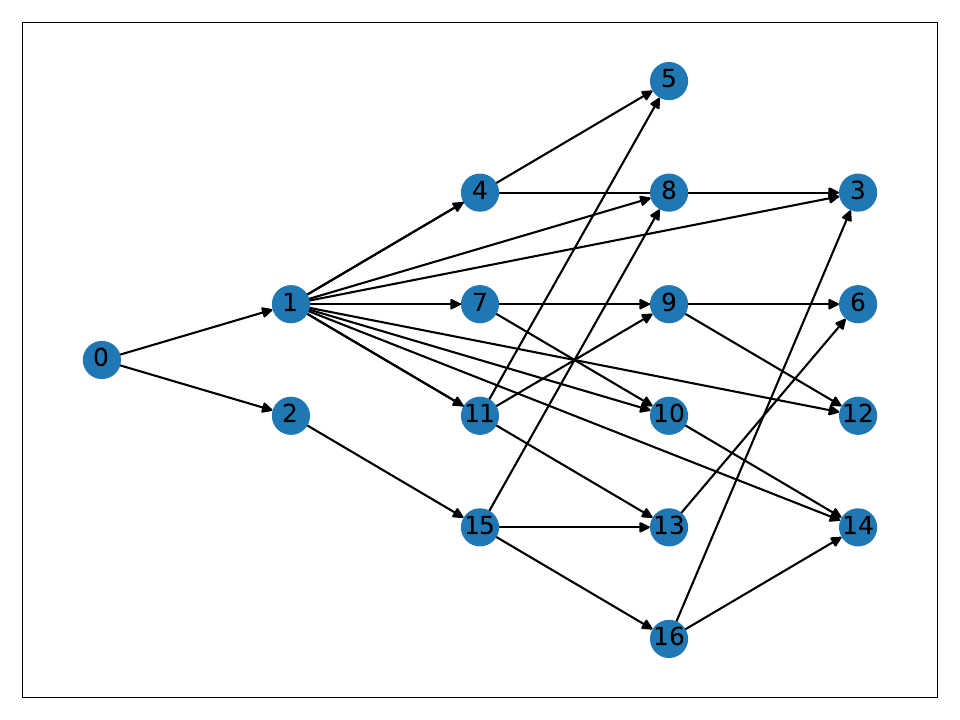}
        \caption{Topography of ``Mother's Delight": A visual representation of the structural segments within this Irish folk tune, illustrating an overall division into A and B sections.}
        \label{fig:trip}
    \end{minipage}\hfill
    \begin{minipage}{0.46\textwidth}
        \centering
        \includegraphics[width=1\linewidth]{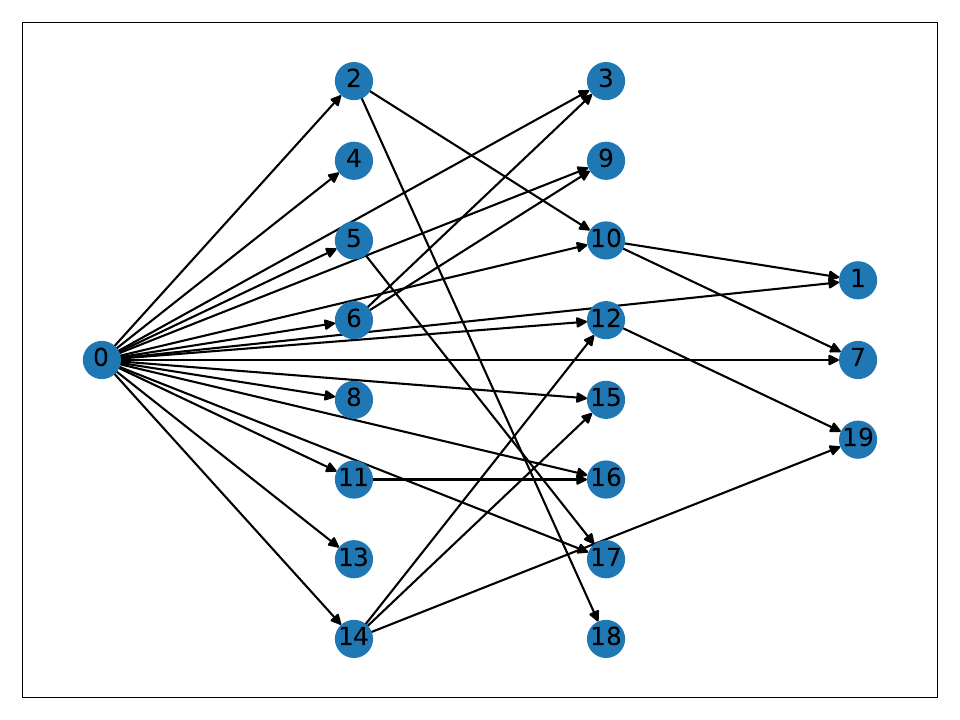}
        \caption{"Hazin" from the Iranian classical corpus: Showcasing the interconnectedness of musical ideas without distinct large sections, highlighting the non-metric nature of Avaz.}
        \label{fig:hazin}
    \end{minipage}
\end{figure}

As we mentioned in the introduction, middle eastern non-metric music necessitates a micro-level analysis rather than a bar-centred analysis which can be suitable for Western compositions. Figure~\ref{fig:trip} illustrates the topology of ``Mother's Delight," an Irish tune. In this topography, nodes represent the rules in the grammar for that piece of music, and the arrows branching from a node indicate other rules that are part of that rule's definition. where '0' signifies the entire tune, and `1' and `2' mark two principal sections, comparable to what might be termed `A' and `B' sections in the Irish tradition. Such sectional divisions are common in Western music.

Conversely, Figure~\ref{fig:hazin} depicts ``Hazin'', one of the Gushehs from our Iranian corpus, where `0' — the entire tune — interlinks directly with eight nodes, indicating the absence of distinct large sections. This structural distinction is crucial; we know from previous work that our grammatical mutation method produces results with structures more similar to that shown on the right. This shows us something important: when making new versions of tunes, we don't need to worry about following strict, large-scale structures. Our method for creating new music works well because it doesn't rely on these large structures either. It can make new and interesting versions of tunes, which is perfect for Avaz or other similar music styles. It suggests our algorithm is a good fit for creating new variations of these tunes.

\subsection{Statistical Results}
\begin{figure}
  \centering
  \includegraphics[trim={0 1cm 0 0}, clip, width=1\textwidth]{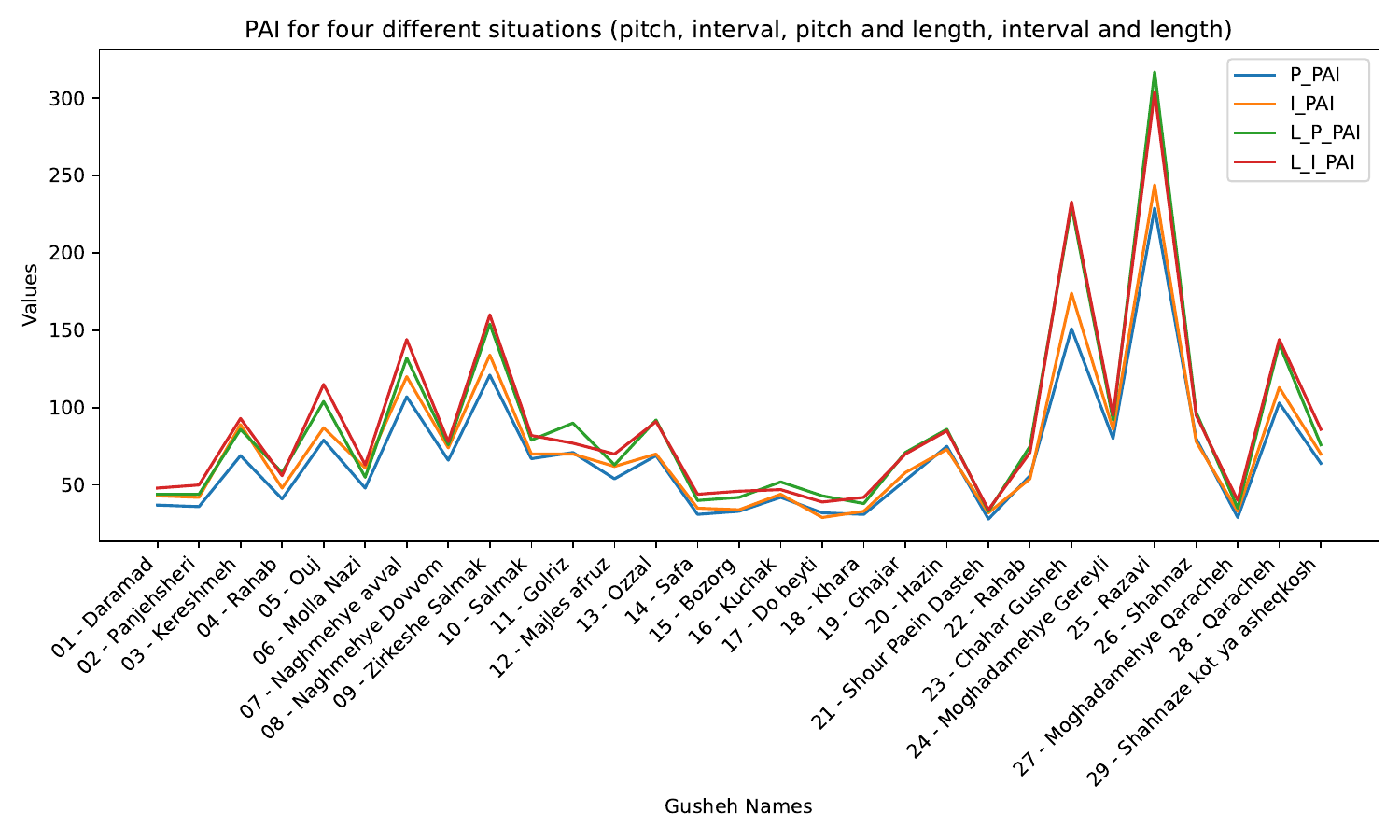}
  \caption{PAI for all members of the corpus based on pitch (P\_PAI), interval (I\_PAI), pitch-and-length tuples (L\_P\_PAI) and interval-and-length tuples (L\_I\_PAI)\label{fig:PAI}}
\end{figure}

We calculated the PAI for all tunes in the corpus, as depicted in Figure~\ref{fig:PAI}. Longer tunes tend to have higher PAI values. The PAI calculation was performed on 4 representations of each tune 1) the pitch, 2) the interval, 3) pitch and length tuples 4) interval and length tuples. Typically, PAIs based solely on pitch are lower compared to those incorporating intervals or length.

Figure~\ref{fig:statistic} presents our findings from applying different metrics to the five setups previously outlined(pitches-chromatic-forward-tuple-non-repair (setup\_1), intervals-chromatic-forward-tuple-clamp (setup\_2), intervals-chromatic-backward-tuple-clamp (setup\_3), intervals-diatonic-backward-tuple-clamp (setup\_4) and intervals-diatonic-backward-tuple-mirror (setup\_5)). This analysis encompasses all Gushehs in our corpus, with each undergoing 100 mutations, represented by individual thin lines. The thick line illustrates the aggregate average across all mutations. We focused on different metrics for this evaluation:

\begin{figure}

  \centering
  \includegraphics[width=0.8\textwidth]{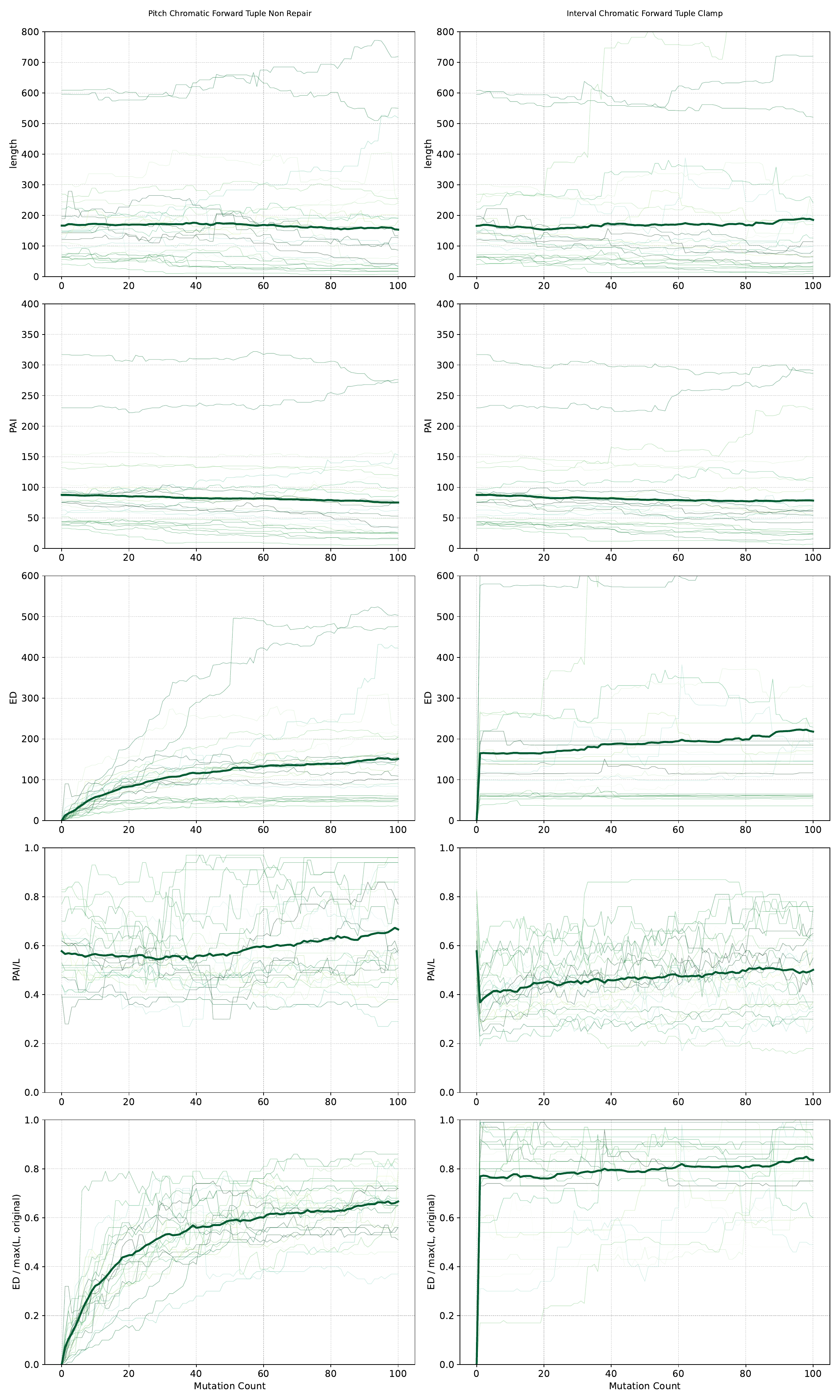}
  \caption{Comparative analysis of two representations across metrics such as length, Edit Distance (ED), Pathway Assembly Index (PAI), and their normalized variants, showing the impact of 100 mutations per Gusheh. Each thin line represents 100 successive mutations on a single Gusheh. The thick line in each plots represents the average, illustrating trends in musical complexity and structural changes.\label{fig:statistic}}
\end{figure}

\begin{figure}

  \centering
  \includegraphics[width=1\textwidth]{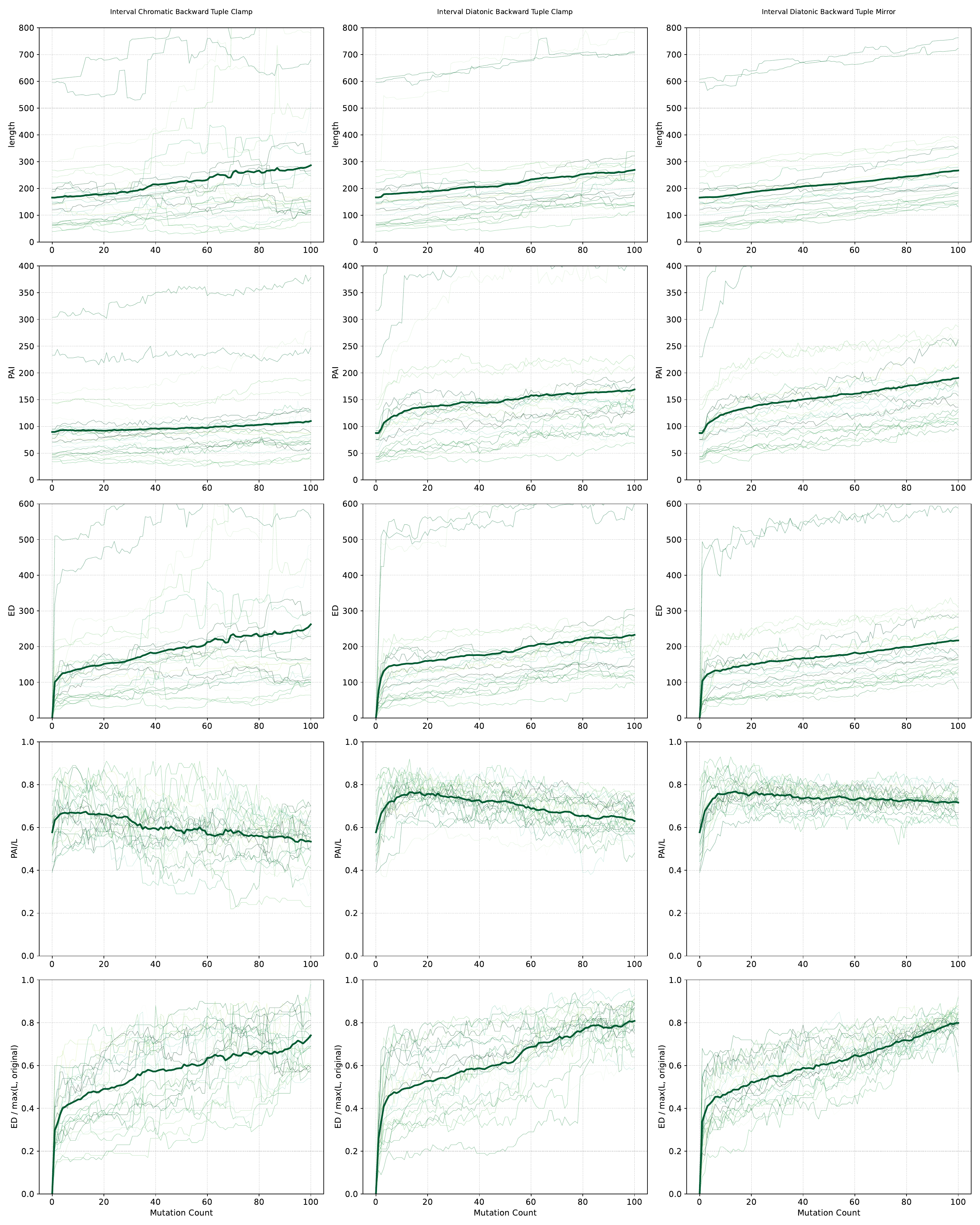}
  \caption{Continuation of Fig.~\ref{fig:statistic}, for the three remaining representations.}
\end{figure}

\textbf{Length}: Shows how the lengths of the Gushehs have been changed by mutations.

\textbf{Edit Distance (ED)}: ED quantifies the changes a Gusheh undergoes after each mutation, comparing each version to the original. An initial ED of 0 indicates no change, as the original tune and itself are identical. A higher ED value per step signifies more substantial modifications, while a gradual increase in ED may be preferred for minor adjustments.

\textbf{Pathway Assembly Index (PAI)}: PAI measures the complexity of the music from the original tune, where closeness to the original composition is considered advantageous. It is a common observation that music often occupies a ``middle ground'' or ``goldilocks zone'': it should not be too simple, nor too complex~\citep{mccormack2021enigma}. The ideal complexity can be estimated by the distribution of complexity values observed in the original corpus. Therefore, if our mutations are successful, they should give results in the same range.  

\textbf{Normalized PAI}: Given that longer tunes naturally exhibit higher PAI values, we normalized these figures by dividing PAI by the tune length (PAI/L) to compare tunes of varying lengths on an equal footing.

\textbf{Normalized ED}: For similar reasons, we normalized ED by dividing it by the maximum length between the mutated tune and the original (ED/ max(L, original)), again enabling a proportionate comparison of changes across tunes of different lengths.


Upon reviewing the results displayed in Figure~\ref{fig:statistic}, derived from the examination of all Gushehs in our corpus across the five setups, we observe the following trends based on 100 mutations per Gusheh:

Keeping the length around the original length is desirable, and none of the setups have a significant change in their length. The PAI values across all setups did not undergo large changes. PAI/L exhibits interesting behavior across the setups. For setup\_1, it gradually increases, indicating a steady rise in complexity relative to length. Setup\_2 had a drop on PAI/L and then started to slightly increase over mutations. In the case of setup\_3, there is an initial increase followed by a decrease. Both setup\_4 and setup\_5 show an increase followed by stability, indicating that after reaching a certain level of complexity, the mutations no longer significantly alter the tune's complexity relative to its length. ED changing behavior is generally similar across all setups. There is a very fast increase in the early mutations and then a slow increase. 




In summary, setup\_1 and setup\_2 are more consistent in maintaining the original length of tunes. setup\_1's PAI values remain closer to the original, suggesting a lesser degree of complexity change. While the interval-based versions reach higher ED values sooner, it does not necessarily mean they offer the desired change more quickly. These setups should be audibly evaluated to determine if the tunes remain recognizable after mutations. The PAI/L metric suggests that all setups maintain the complexity of the music well except setup\_2, indicating that each possesses valuable features for mutation. 

\subsection{Musical Observations}

In conducting the analysis presented within this section, it is important to comment the viewpoint brought by the first author, who is a classically-trained Iranian musician. While the insights derived are deeply informed, they are subjective and influenced by the individual’s specific training. We are in the process of designing more objective listening tests for future work.

Our evaluation involved listening to variations of Gushehs after 5, 10, 15, and 20 mutations across the entire corpus, utilizing the five setups introduced earlier. We concentrated on mid-sized Gushehs, avoiding extremely long or short pieces.

\textbf{Early Mutations (5 and 10)}: After 5 mutations, pitch-based variations remained closely recognizable compared to their original counterparts, with changes being noticeable yet coherent. The interval-based versions, while sometimes close to the original tune, often resulted in more pronounced alterations. By the 10th mutation, pitch-based adaptations provided a musically suitable (based on the author's experience) new version of the Gushehs, while maintaining the essence of Gushehs. However, jumps that are not regular in this style of music were observed, attributed to a broader modal framework in some tunes, which is less typical in traditional Iranian classical music. Interval-based versions at this stage diverged more significantly, offering a distinctly different interpretation of the original Gusheh.

\textbf{Mid to Late Mutations (15 and 20)}: The transition through the 15th mutation did not introduce major changes, serving more as an intermediate step. Upon reaching 20 mutations, pitch-based versions evolved into distinctly different interpretations, suggesting potential avenues for novel improvisation. Yet, they retained musical integrity, suggesting creativity within bounds. Conversely, 20 mutations in interval-based setups often resulted in confusion, with changes deemed excessive for considering them as mere variations. Among the interval-based approaches, the mirror repair setup frequently yielded the most musically pleasing outcomes, suggesting its effectiveness in preserving musicality while introducing novelty.

\textbf{Pitch vs. Interval-Based Mutations}: Pitch-based mutations generally led to more musically coherent variations, possibly due to their ability to retain recognizable patterns. Interval-based mutations, while maintaining patterns, shifted them in ways that could start from unexpected notes, leading to confusion regarding the original tune’s essence.

\textbf{Special Note on Alterations}: Mutations involving altered notes, such as altered notes in Salmak, occasionally produced dissonant sounds. That happens notably on the fifth degree of the tonal centre, which in our dataset is D. Bringing D, then some other notes, and then Dk is common in Iranian classical music, but the reverse order is not and results in a dissonant interval. Such alterations clashed with the traditional rules of Iranian classical music, highlighting areas where the algorithm might benefit from incorporating musical context and rules to avoid unintended dissonance.

In summary, our listening experience revealed that pitch-based approach generally offered more suitable variations.

\subsection{Algorithm results on Concatenated Dastgah}

 One of the main objectives when a musician plays the entire concatenated Dastgah is to preserve Seyr, or melodic movement~\citep{chalesh2017seyir}. In the context of traditional Iranian music, \textit{Seyr} or``melodic movement'' encapsulates the progression of melodies within a piece, guiding the overall direction of the melody through its initiation, development, climax, and resolution. Observing the order of introducing frameworks is one of the main features of Seyr. Seyr emphasizes the importance of transitional notes and melodic phrases in establishing the identity and modal character of the piece. These elements are crucial in directing the melodic flow from one segment to another, ensuring a coherent and expressive musical journey. Here, we carry out a new experiment on the concatenated Dastgah.
 
\begin{figure}
    \centering
    \includegraphics[width=1\textwidth]{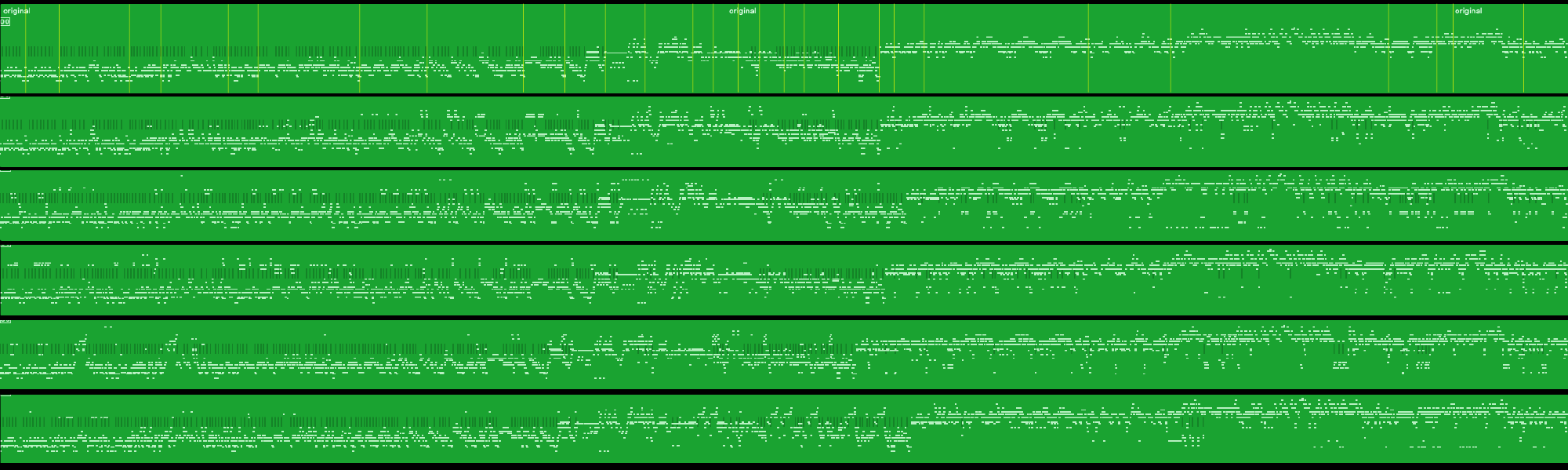}
    \caption{ After making variations, Seyr remains very similar to its original version. \label{fig:entire}}
\end{figure}

In our corpus, PAI is 1982 for the sum of individual PAIs, while that is 1406 for considering the Dastgah as one piece of music. This is a common principle in MDL. If there are any commonalities across the corpus, the compressed description of the concatenated corpus will be less than the sum of compressed descriptions of the individual parts. The early parts of the piece can provide a “dictionary” for later parts, which allows savings. In the first row of Fig.~\ref{fig:entire} we can see the entire Dastgah, each Gusheh is shown with yellow highlights. The trend of fluctuation in this row shows us the Seyr for the original version.

By comparing MIDI files shapes in Fig.~\ref{fig:entire} it can be seen how in these five variations the Seyr has been well-preserved compared to the original one. In fact, each piece usually stays in the same modal framework, even after many mutations. When a Gusheh changes the previous modal framework, the introduced framework remains prominent in the new version. As a result, the overall trend (Seyr) remains similar. PAI for these variations respectively are 1555, 1517, 1555, 1557, 1572 which shows around 8 to 11 percent increase in PAI after 100 mutations. Together with our results in the previous subsection, this suggests that our mutation methods have a slight bias towards increasing complexity and/or length.

\subsection{More Applications of Parsing}
The parsing method offers potential for educational and (ethno)musicological applications. The traditional process of teaching Radif emphasizes learning by listening, where teachers break down Gushehs into small, manageable pieces for students to easily grasp and replicate. This approach relies heavily on repetitive segments, which, as shown in Figure~\ref{fig:transcript}, the algorithm in most cases accurately identifies. We highlight only the initial occurrence of each segment, but the tune's grammar reveals their repetition and structure~\ref{fig:Grammar}. Such grammatical analysis could be invaluable in educational settings, aiding students in learning non-metric music through its conventional methods.

A crucial aspect of Radif performance is recognizing significant motifs within the repertoire. Ali Jariteh's manual analysis of the entire Radif~\citep{Jari2023} to identify these motifs or phrases is an attempt to find and introduce these motifs and phrases. The algorithm can help with this kind of research, for instance comparing these kind of motifs in two different versions of Radif, as the algorithm is a tool to discover these patterns. While Figure~\ref{fig:transcript} displays motifs identified by the algorithm within a single Gusheh, running the algorithm across an entire Dastgah uncovers longer phrases repeated across multiple Gushehs. An example is a cadence phrase in line 8, echoing similar phrases in other tunes, or the sequence in lines 4, 5, and 6, which resembles the first tune in our corpus (Daramad). By detecting these similarities, the algorithm simplifies understanding the connections and relationships between Gushehs.

\section{Conclusion and Future Works}

In this work, we introduced a symbolic corpus for Iranian classical non-metric music. Furthermore, we explored different representations for this style of music and adapted the PA algorithm to the corpus to find its underlying hierarchical structure. We also used the algorithm to generate musical variations based on five different setups. After conducting auditory and statistical analyses, it turns out that pitch-based variations become more musically suitable after about 10 mutations. Future work will focus on developing a system to identify the most suitable mix of setups for a specific tune, aiming to generate variations. We have released all of our data, but in future we will expand our dataset to include the entire Radif book.

\section{Acknowledgments}

This work was conducted with the financial support of the Science Foundation Ireland Centre for Research Training in Digitally-Enhanced Reality (d-real) under Grant No. 18/CRT/6224.

\section{Ethics Statement}

This study centers on the exploration of algorithmic variations in non-metric Iranian classical music. Our work uses publicly available music corpora and does not involve human participants directly. The music collection is part of the Iranian classical music tradition, which is not itself owned or copyrighted by any individual composer. The edition we have used is used by permission of the author.

This musical tradition is culturally important, and we have tried to be respectful and careful in how we talk about it and use it.

The authors declare no conflicts of interest.

\bibliographystyle{plain}
\bibliography{refs}

\begin{thebibliography}{10}

\bibitem{almazaydeh2022arabic}
Laiali Almazaydeh, Saleh Atiewi, Arar Al~Tawil, and Khaled Elleithy.
\newblock Arabic music genre classification using deep convolutional neural networks (cnns).
\newblock {\em Computers, Materials \& Continua}, 72(3), 2022.

\bibitem{baba2019nava}
B~Baba~Ali, A~Gorgan~Mohammadi, and A~Faraji~Dizaji.
\newblock Nava: A persian traditional music database for the dastgah and instrument recognition tasks.
\newblock {\em Advanced Signal Processing}, 3(2):125--134, 2019.

\bibitem{chalesh2017seyir}
Maryam Chalesh and Hooman Asa’di.
\newblock “seyir” in persian classical music (case study: Dar{\^a}mad-e av{\^a}z-e bay{\^a}te-esfah{\^a}n).
\newblock {\em Journal of Fine Arts: Performing Arts \& Music}, 22(2):81--91, 2017.

\bibitem{doherty2022melodic}
Se{\'a}n Doherty.
\newblock Melodic structures in the double jigs of o’neill’s the dance music of ireland: 1001 gems (1907).
\newblock {\em Journal of the Society for Musicology in Ireland}, pages 19--45, 2022.

\bibitem{during1997systeme}
Jean During.
\newblock Le syst{\`e}me des offrandes dans la tradition ahl-e haqq.
\newblock In {\em Syncretistic Religious Communities in the Near East}, pages 49--64. Brill, 1997.

\bibitem{during2006intervals}
Jean During.
\newblock The intervals of the azerbaijani mugam: Back to the sources.
\newblock In {\em international symposium on Turkish Ottoman period. Bursa. April}, 2006.

\bibitem{during1991art}
Jean During, Zia Mirabdobaghi, and Dariush Safvat.
\newblock The art of persian music.
\newblock {\em (No Title)}, 1991.

\bibitem{farhat1978form}
Hormoz Farhat.
\newblock Form and style in persian music.
\newblock {\em The World of Music}, 20(2):109--118, 1978.

\bibitem{farhat2004dastgah}
Hormoz Farhat.
\newblock {\em The dastgah concept in Persian music}.
\newblock Cambridge University Press, 2004.

\bibitem{grunwald2007minimum}
Peter~D Gr{\"u}nwald.
\newblock {\em The minimum description length principle}.
\newblock MIT press, 2007.

\bibitem{heydarian2005database}
Peyman Heydarian and Joshua~D Reiss.
\newblock A database for persian music.
\newblock In {\em Proc. of the Digital Music Research Network Summer Conference (DMRN 2005)}, 2005.

\bibitem{Jari2023}
Ali Jari.
\newblock Analysis of melodic figures in mirza abdullah's radif.
\newblock {\em Mahoor Music Quarterly}, 46(Winter 2009):75--121, 2009.
\newblock 224 pages, Cover price: 60000 IRR, Published on 2010-03-16.

\bibitem{Kanani2018}
M.~Kanani.
\newblock {\em Taknavazi Tar Mahoor}.
\newblock Narvan, Tehran, 1st edition, 2018.
\newblock Yahya Zarpanje.

\bibitem{kanani2019role}
Maziar Kanani and Hooman Asadi.
\newblock The role of tombak as an accompaniment percussion in iranian classical music (1946-2006).
\newblock {\em Honar--Ha-Ye-Ziba: Honar-Ha-Ye-Namayeshi Va Mosighi}, 24(1):5--14, 2019.

\bibitem{kanani2020study}
Maziar Kanani and Mohammad~Reza Azadehfar.
\newblock Study of tahrir patterns in iranian classical radif music based on mohammadreza shajarian's performance.
\newblock 2020.

\bibitem{kanani2023graph}
Maziar Kanani, Se{\'a}n O'Leary, and James McDermott.
\newblock Graph-based mutations for music generation.
\newblock In {\em Proceedings of the Companion Conference on Genetic and Evolutionary Computation}, pages 1916--1919, 2023.

\bibitem{kanani2023parsing}
Maziar Kanani, Se{\'a}n O’Leary, and James McDermott.
\newblock Parsing musical structure to enable meaningful variations.
\newblock In {\em AIMC 2023 (forthcoming): The International Conference on AI and Musical Creativity, Sussex, UK, 30th August-1st September}, 2023.

\bibitem{karaosmanouglu2012turkish}
M~Kemal Karaosmano{\u{g}}lu.
\newblock A turkish makam music symbolic database for music information retrieval: Symbtr.
\newblock In {\em Proceedings of 13th International Society for Music Information Retrieval Conference; 2012 October 8-12; Porto, Portugal. Porto: ISMIR, 2012. p. 223--228.} International Society for Music Information Retrieval (ISMIR), 2012.

\bibitem{li2008introduction}
Ming Li, Paul Vit{\'a}nyi, et~al.
\newblock {\em An introduction to Kolmogorov complexity and its applications}, volume~3.
\newblock Springer, 2008.

\bibitem{marshall2019quantifying}
Stuart~M Marshall, Douglas Moore, Alastair~RG Murray, Sara~I Walker, and Leroy Cronin.
\newblock Quantifying the pathways to life using assembly spaces.
\newblock {\em arXiv preprint arXiv:1907.04649}, 2019.

\bibitem{mccormack2021enigma}
Jon McCormack, Camilo Cruz~Gambardella, and Andy Lomas.
\newblock The enigma of complexity.
\newblock In {\em Artificial Intelligence in Music, Sound, Art and Design: 10th International Conference, EvoMUSART 2021, Held as Part of EvoStar 2021, Virtual Event, April 7--9, 2021, Proceedings 10}, pages 203--217. Springer, 2021.

\bibitem{nevill1997identifying}
Craig~G Nevill-Manning and Ian~H Witten.
\newblock Identifying hierarchical structure in sequences: A linear-time algorithm.
\newblock {\em Journal of Artificial Intelligence Research}, 7:67--82, 1997.

\bibitem{nikzat2022kdc}
Babak Nikzat, Kunstuniversit{\"a}t Graz, and Rafael~Caro Repetto.
\newblock Kdc: An open corpus for computational research of dastg {\.z}ahi music.
\newblock 2022.

\bibitem{nooshin2017iranian}
Laudan Nooshin.
\newblock {\em Iranian classical music: The discourses and practice of creativity}.
\newblock Routledge, 2017.

\bibitem{nooshin2006improvisation}
Laudan Nooshin and Richard Widdess.
\newblock Improvisation in iranian and indian music.
\newblock {\em Journal of the Indian Musicological Society}, 36:104--119, 2006.

\bibitem{parlak2018automatic}
{\.I}smail~Hakk{\i} Parlak and Cem K{\"o}semen.
\newblock Automatic music generation by true random numbers for turkish makams.
\newblock In {\em 2018 4th International Conference on Computer and Technology Applications (ICCTA)}, pages 64--68. Ieee, 2018.

\bibitem{schmidhuber1997low}
J{\"u}rgen Schmidhuber.
\newblock Low-complexity art.
\newblock {\em Leonardo}, 30(2):97--103, 1997.

\bibitem{sertan2011modeling}
S~Sertan and Parag Chordia.
\newblock Modeling melodic improvisation in turkish folk music using variable-length markov models.
\newblock In {\em 12th International Society for Music Information Retrieval Conference}, pages 269--274, 2011.

\bibitem{talai2015}
Dariush Talai.
\newblock {\em Radif Analysis Based on the Notation of Mirza Abdollah's Radif With Annotated Visual Description}.
\newblock Ney, Tehran, Iran, 2015.

\bibitem{tsuge1970rhythmic}
Gen'ichi Tsuge.
\newblock Rhythmic aspects of the {\^a}v{\^a}z in {P}ersian music.
\newblock {\em Ethnomusicology}, pages 205--227, 1970.

\bibitem{uyar2014corpus}
Burak Uyar, Hasan~Sercan Atli, Sertan {\c{S}}ent{\"u}rk, Bar{\i}{\c{s}} Bozkurt, and Xavier Serra.
\newblock A corpus for computational research of turkish makam music.
\newblock In {\em Proceedings of the 1st International Workshop on Digital Libraries for Musicology}, pages 1--7, 2014.

\end{thebibliography}
\end{document}